\documentclass[sigconf]{acmart}

\usepackage[subject={Todo},author={Alex}]{pdfcomment}
\usepackage{xcolor}
\usepackage[normalem]{ulem}
\usepackage[textsize=tiny]{todonotes}

\usepackage{float}
\usepackage{placeins}
\usepackage{algorithm}
\usepackage{algorithmicx}
\usepackage{algpseudocode}
\usepackage{colortbl}
\usepackage{fixltx2e}
\usepackage{dblfloatfix}
\usepackage{url}

\usepackage{booktabs} 

\setcopyright{rightsretained}

\acmDOI{10.475/123_4}
\acmISBN{123-4567-24-567/08/06}
\acmConference[GECCO '17]{the Genetic and Evolutionary Computation Conference 2017}{July 15--19, 2017}{Berlin, Germany}
\acmYear{2017}
\copyrightyear{2017}
\acmPrice{15.00}

\begin{document}
\title{Evolving Parsimonious Networks by Mixing Activation Functions}
\subtitlenote{The full version of the author's guide is available as
  \texttt{acmart.pdf} document}

\author{Alexander Hagg}
\orcid{1234-5678-9012}
\affiliation{%
  \institution{Bonn-Rhein-Sieg University of Applied Sciences}
  \streetaddress{Grantham-Allee 20}
  \city{Bonn} 
  \state{Germany} 
  \postcode{53757}
}
\email{alexander.hagg@h-brs.de}

\author{Maximilian Mensing}
\orcid{1234-5678-9012}
\affiliation{%
  \institution{Bonn-Rhein-Sieg University of Applied Sciences}
  \streetaddress{Grantham-Allee 20}
  \city{Bonn} 
  \state{Germany} 
  \postcode{53757}
}
\email{max.mensing@smail.inf.h-brs.de}

\author{Alexander Asteroth}
\orcid{1234-5678-9012}
\affiliation{%
  \institution{Bonn-Rhein-Sieg University of Applied Sciences}
  \streetaddress{Grantham-Allee 20}
  \city{Bonn} 
  \state{Germany} 
  \postcode{53757}
}
\email{alexander.asteroth@h-brs.de}


\begin{abstract}
Neuroevolution methods evolve the weights of a neural network, and in some cases the topology, but little work has been done to analyze the effect of evolving the activation functions of individual nodes on network size, which is important when training networks with a small number of samples.
In this work we extend the neuroevolution algorithm NEAT to evolve the activation function of neurons in addition to the topology and weights of the network. The size and performance of networks produced using NEAT with uniform activation in all nodes, or homogenous networks, is compared to networks which contain a mixture of activation functions, or heterogenous networks.
For a number of regression and classification benchmarks it is shown that, (1) qualitatively different activation functions lead to different results in homogeneous networks, (2) the heterogeneous version of NEAT is able to select well performing activation functions, (3) producing heterogeneous networks that are significantly smaller than homogeneous networks.
\end{abstract}

%
%




\maketitle

\section{Introduction}
\label{sec:intro}
In many areas of interest for machine learning, sample sets are small and noisy, for example in robotics where environments are prone to continuous changes and novelty or when using machine learning models to replace expensive sampling methods in optimization. These complex problems often require models with a high number of degrees of freedom, which tend to overfit when presented with only a small number of samples.

Neuroevolutionary (NE) methods represent a starkly different neural network training paradigm than backpropagation of error. Morse and Stanley~\cite{Morse2016} point out that a number of advantages, like the regularizing properties of topological evolution~\cite{Stanley2002a}, diversity maintenance and indirect encoding, allows NE to perform certain tasks better. Especially when using NE for regression, where models need to be created that are as general as possible, NE training methods allow to find simple/minimal models, adhering to the law of parsimony. Smaller networks have been shown to be less prone to overfitting~\cite{orr1993}, and are easier to train, because less model parameters have to be optimized. 

To understand how we can improve NE methods further, a number of questions still needs to be answered. Taking advantage of the strengths of NE, for example by increasing the expressiveness of networks with a higher diversity of neural activation functions, which is harder for classical methods, we could decrease the network size and limit overfitting. NE could allow training neural networks for small sample sets which are too small to train large networks, by topology minimization and decrease overfitting for problems that are hard to represent with small networks.
 
In this work, we try to answer the first question and look at the evolution of the activation functions in a topology-evolving algorithm and its effect on network size, convergence speed and error approximation in a number of benchmark regression problems.

\subsection{Influence of Activation Function}

Hornik showed that the Universal Approximation Theorem\footnote{Based on Kolmogorov's superposition theorem~\cite{kolmogorov22representation}} is valid for neural networks using \textit{any} continuous nonconstant function, stating that for these activation functions, multilayer feedforward neural networks are capable of arbitrarily accurate approximation to any real-valued continuous function, as long as outputs are bounded~\cite{Hornik1989}.

\begin{figure}[h!]
\includegraphics[width=\linewidth]{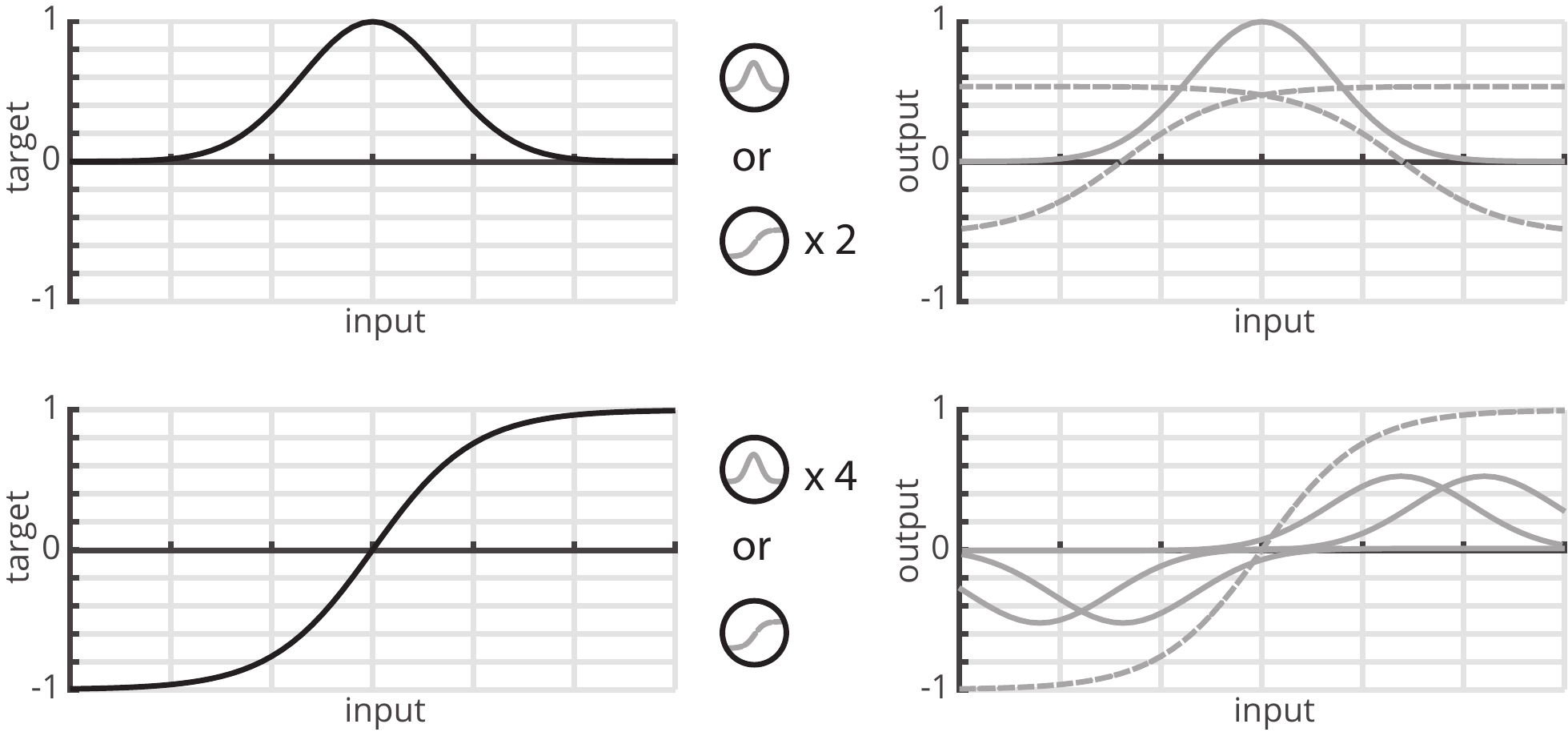}
\caption{Left: target functions (Gaussian and sigmoid function). Middle: how many neurons are theoretically necessary to represent the target with a Gaussian and sigmoid activation function. Right: shows how these neurons need to be positioned and scaled to approximate the target.}
\label{fig:howmanyneurons}
\end{figure}

Although the choice of an activation function does not have an influence on the \textit{theoretical expressiveness} of neural networks, it can affect the \textit{training behaviour} of the network. The number of nodes needed to approximate a given function is highly dependent on the activation function of the nodes. To approximate a Gaussian function two sigmoid functions are required, to approximate a sigmoid function four Gaussian functions are needed (Figure \ref{fig:howmanyneurons}).

Although this example is contrived, it is easy to see that the number of nodes needed is dependent on the used activation function. As more nodes are required, the number of weights that need to be trained also grows, this expands the search space of the training algorithm, and extends convergence times. Several studies have shown the significant influence the choice of activation function can have on the performance of neural networks. Kamruzzaman and Aziz~\cite{Kamruzzaman2002} show that activation functions influence convergence speed, depending on the target function. In a large literature review by Laudani et al~\cite{Laudani2015} comparing many activation functions, the authors find large differences in terms of convergence speed (in epochs) and resulting network errors. An extensive comparison was done by Efe~\cite{Efe2008}, using 8 benchmark data sets, drawing the same conclusion, that different activation functions result in different convergence behavior and accuracy, with respect to the resulting network errors.

\subsection{Heterogeneous Networks}
\label{sec:intro:heterogeneous}
\begin{figure}[h!]
\includegraphics[width=\linewidth]{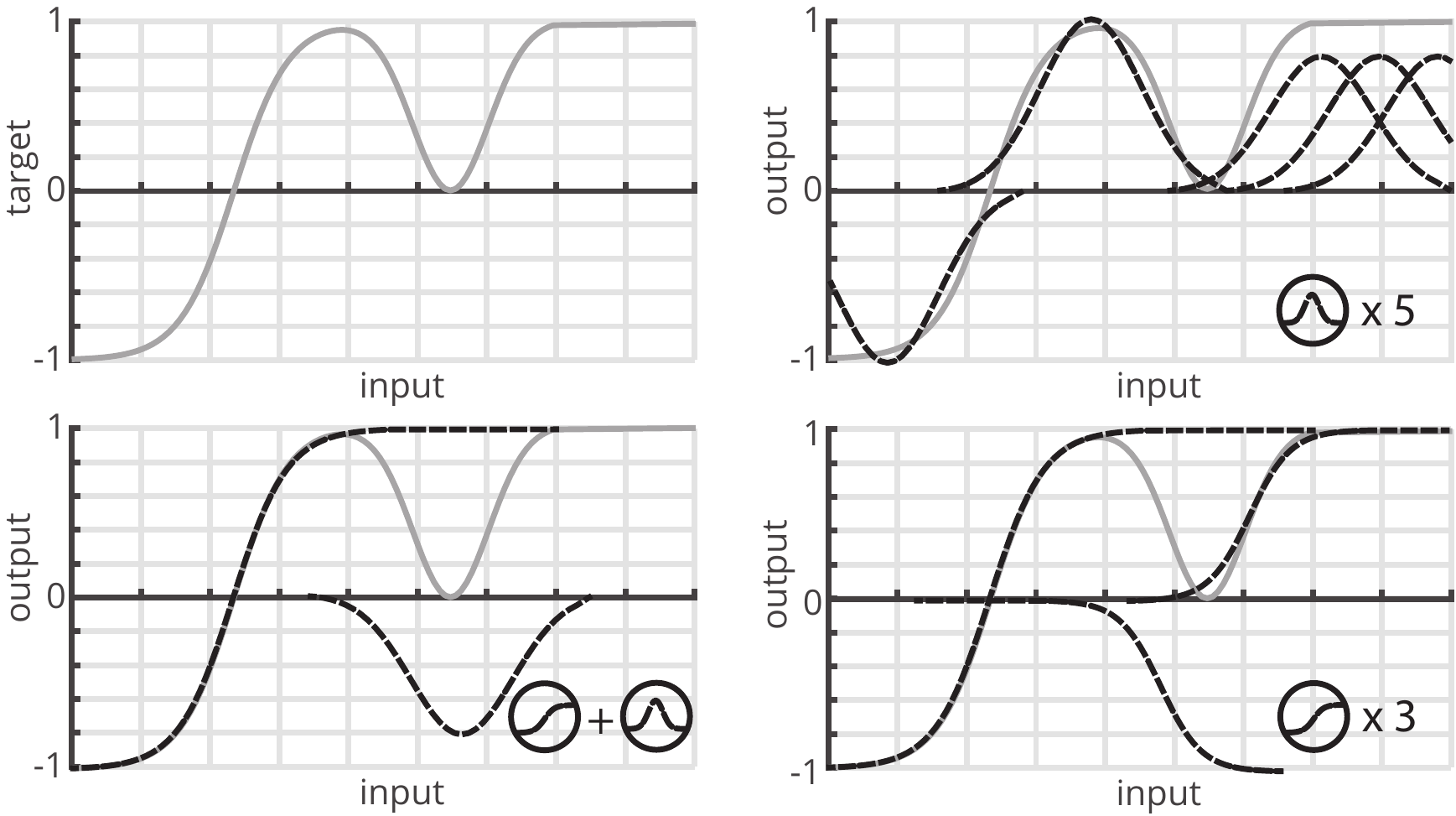}
\caption{Target function combined from a Gaussian and sigmoid. In this case, two neurons (bottom left), one with a Gaussian and one with a sigmoid activation, would suffice to accurately predict the target function. Homogeneous networks that only use a single activation function (right) need more neurons to approximate the same target.}
\label{fig:heteroactivation}
\end{figure}

In contrast to backpropagation, the training method in NE is independent of function derivatives. Since the best non-linearity is unknown beforehand, we can easily use NE training to create networks containing a heterogeneous set of activation functions. Figure \ref{fig:heteroactivation} shows an example where it makes sense to use such networks. In this case, combining a Gaussian and a sigmoidal activated neuron leads to a good approximation, whereas a homogeneous network, which uses only one activation function, would need more neurons and weights, increasing the size of the topology-weight search space.

\begin{figure}[h!]
\includegraphics[width=\linewidth]{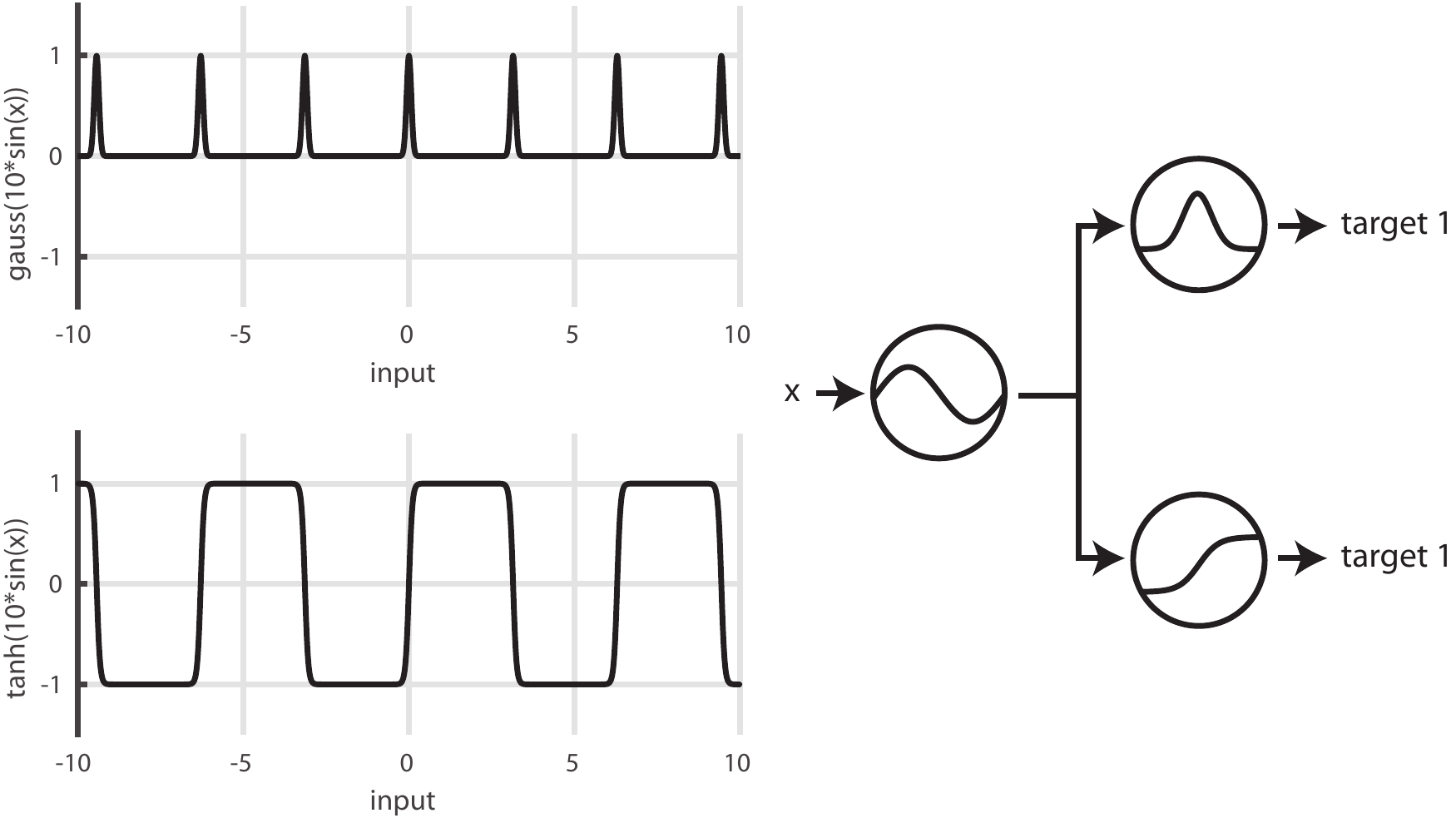}
\caption{Multiple activation functions allow for correlated targets to be learned with less nodes. Here, a sine function is used in the first layer, after which a Gaussian and sigmoid kernel are applied to produce two different output targets, $\sigma$(10*sin(x)) and tanh(10*sin(x)). }
\label{fig:multitarget}
\end{figure}

The outputs in multi-target problems are often correlated. When we try to approximate the vibration frequency and temperature response in a motor to the amount of acceleration in a car, both targets are likely to be correlated by the response of the valves in the motor. These valves will contain a dampening effect, which depends on the input (the gas pedal), which will affect both targets. Models that approximate and predict these target values benefit when they are able to describe this global effect for all targets. Figure \ref{fig:multitarget} shows an example of a heterogeneous neural network approximating two such correlated target functions. The sine function in the first neuron describes a part of both targets. The second layer of neurons can then specialize into learning decorrelated signals. This modularity is something we expect to see even in small heterogeneous networks and is another reason why the use of heterogeneous networks can lead to more parsimonous models.

Not using a fixed activation function and instead developing it with the network's topology allows for compact networks to be more expressive and decrease the size of the parameter space. Not only are smaller networks more easily trained, but that training can be performed with fewer samples, and the resulting networks are less prone to overfitting of the training data~\cite{orr1993}. Especially in cases where samples are very noisy or the number of available is not sufficient to train large networks, this can be an advantage compared to classical training methods.

There is no analytical way to adapt the activation function in on-line backpropagation learning and often, model selection methods like the Akaike Information Criterion \cite{Angus1991} need to be used to select the appropriate model. Neuroevolution offers methods that allow coevolution of activation functions in an on-line manner. We extend the NEAT~\cite{Stanley2002a} algorithm and refer to it as Heterogeneous Activation NEAT (HA-NEAT), allowing it to evolve the activation function as well, using a similar idea as was applied in HyperNEAT~\cite{Stanley2009}, but producing directly encoded networks. We then compare the performance of the original implementation, producing homogeneous neural topologies, to HA-NEAT. 

\section{Related Work}
\label{sec:related}
\subsection{Evolving Topologies}
\label{sec:related:topologies}
While some neuroevolutionary algorithms like SANE~\cite{Moriarty1996} or ESP~\cite{Gomez1998} evolve only fixed networks, others like NEAT evolve the network's topology as well. The NEAT algorithm was introduced by Stanley and Miikkulainen in 2002~\cite{Stanley2002a}. To keep the produced topologies as small as possible, NEAT follows a bottom up approach, initialising minimal networks in which inputs are connected directly to outputs. Then, slowly, neurons and connections are added by mutation operators, gradually complexifying the networks. While most parts of a neural network are evolved, a fixed activation function is used for all hidden neurons.
NEAT does not suffer from \textit{bloat}, growing network size with little improvement to fitness. This would slow the evolutionary process, increasing the algorithm's memory footprint and hampering breeding methods~\cite{Trujillo2014}.

Stanley et al. extended NEAT to HyperNEAT seven years later~\cite{Stanley2009}. They evolve Compositional Pattern Producing Networks (CPPN)~\cite{Stanley2007} that output a function which determines the weights in a preconfigured neural network which depends on the position of the connection's source and target neuron's coordinates in a predefined coordinate system, thus indirectly encoding the network's weight matrix. CPPNs are evolved as a composite function, which includes multiple activation functions. The algorithm is mostly used to solve control problems~\cite{Taras2014}, where locality or symmetries in the neurons' locations can be defined in a meaningful way. 

Khan et al. introduce an extension to Cartesian Genetic Programming (CGP)~\cite{Miller2000,MahsalKhan2013}, CGPANN, in which they define neurons as nodes in the cartesian genotype. CGP allows neuron connections to "jump" layers and thus allowing topologies to change during evolution. The algorithm does not use "minimal initialization" like NEAT and does not emphasize creating parsimonious networks that generalize well.

\subsection{Evolving Activation Functions}
\label{sec:related:activation}
Neural networks usually utilise the same, non-linear fixed activation function for all neurons, but the activation function has a significant influence on the learning performance, topology and fitness~\cite{Kamruzzaman2002,Efe2008,Laudani2015}.

Mayer and Schwaiger evolve the activation function of generalized multi-layer perceptrons~\cite{Mayer1993} within the netGEN framework, which uses a genetic algorithm to evolve neural network topologies~\cite{Huber1995}. The activation functions are described in the genome by the coordinates of the control points of a cubic spline function. This also allows non-monotonic activation functions. The authors show that by evolving the activation function they are able to learn the XOR problem with a single neuron as well as increase the accuracy of the evolved networks on more complex target functions. The effect on the evolutionary process, on convergence times, is not investigated. Yao gives an overview on the evolution of neural activation functions~\cite{Yao1999} and show that most work was done mixing sigmoidal and Gaussian neurons, and thus selecting the transfer functions from a predefined set.

In recent work, Turner and Miller~\cite{Turner2014} acknowledge that a lot of work has been left open on NE that includes evolving the transfer function. 
They use a single scaling parameter that is learned during training to evolve heterogeneous networks. They show that the effectiveness of using NE with either a convential fixed-topology NE algorithm or Cartesian Genetic Programming (CGP) to train homogeneous ANNs is dependent on the selected activation function. They evaluate their approach using a number of classification benchmarks, reinforcement learning benchmarks and a simple circuit regression task and evolved variable activation functions by including a function scaling factor. The authors show that homogeneous networks reach significantly different fitness levels and convergence times, depending on the activation function and data set, whereby no single activation function can be shown to work best for all problems. Heterogeneous networks are created as well with the transfer function evolved from both a list of the beforementioned nonlinear functions or as a parameterized activation function. The authors find clear improvements using heterogeneous networks, when comparing to the \textit{average} homogeneous network. In most cases however, there exists an activation function for which the homogeneous networks perform mostly better than the heterogeneous ones. They also find that the networks do not tend to prefer any particular activation function, especially in the case of the CGP algorithm that evolves the topology as well as the weights. 

\subsection{Discussion}
\label{sec:related:discussion}
It is necessary to do more research on the effects of evolving the transfer functions in neuroevolved networks, especially in the context of regression and algorithms that construct a network's topology.

As NEAT is a NE method that initializes networks in a minimal state, which will allow us to use this regularizing effect, this work extends that approach. The algorithm is naturally topology-minimizing and its internal representation is easily extendible to include coevolution of activation functions.
Especially when targeting regression where multiple target outputs are correlated, the bottom-up construction of neural networks in NEAT could allow the more general submodels or \textit{modules} to evolve first, after which the later layers or neurons are added to specialize on more specific behaviours in the underlying model.

The original implementation of NEAT does not evolve the activation functions of the network, leading to homogeneous networks. Even though there is research on evolving activation functions~\cite{Mayer1993,Gauci2010,Agostinelli2014}, these methods have not yet been applied to the canonical NEAT. While HyperNEAT, a NEAT derivate, evolves activation functions, it has many other modifications. Especially noteworthy is the use of CPPNs instead of the directly encoded ANNs NEAT uses~\cite{Gauci2010,Risi2010}. 
While this approach increases the suitability to solve high-dimensional control and vision problems, apart from symbolic regression~\cite{Drchal2012} it has not been widely applied on regression or classification problems. HyperNEAT and its derivatives were designed to indirectly train large networks, but we focus on training directly encoded parsimonious networks for regression.

\section{Approach}
\label{sec:approach}
In order to grow heterogeneous models, we enhance NEAT by allowing the evolutionary process to affect neurons' activation functions. This enhanced version of the algorithm, HA-NEAT, produces heterogeneous topologies, increasing the parsimony of NEAT, shortening the time to convergence, and increasing the accuracy of the produced model. The rest of this section only discusses changes we made to the original implementation of NEAT. Please refer to the original publication for elaborated details on the implementation of NEAT~\cite{Stanley2002a}. 

To simplify optimization of topology, and because of our focus on regression problems, we only allow feed forward networks. Before a connection is added to the network, we check whether it creates a cycle in the network. Only connections that do not cause such a recurrency are added.

\begin{figure}[ht!]
\centering
\includegraphics[width=\linewidth]{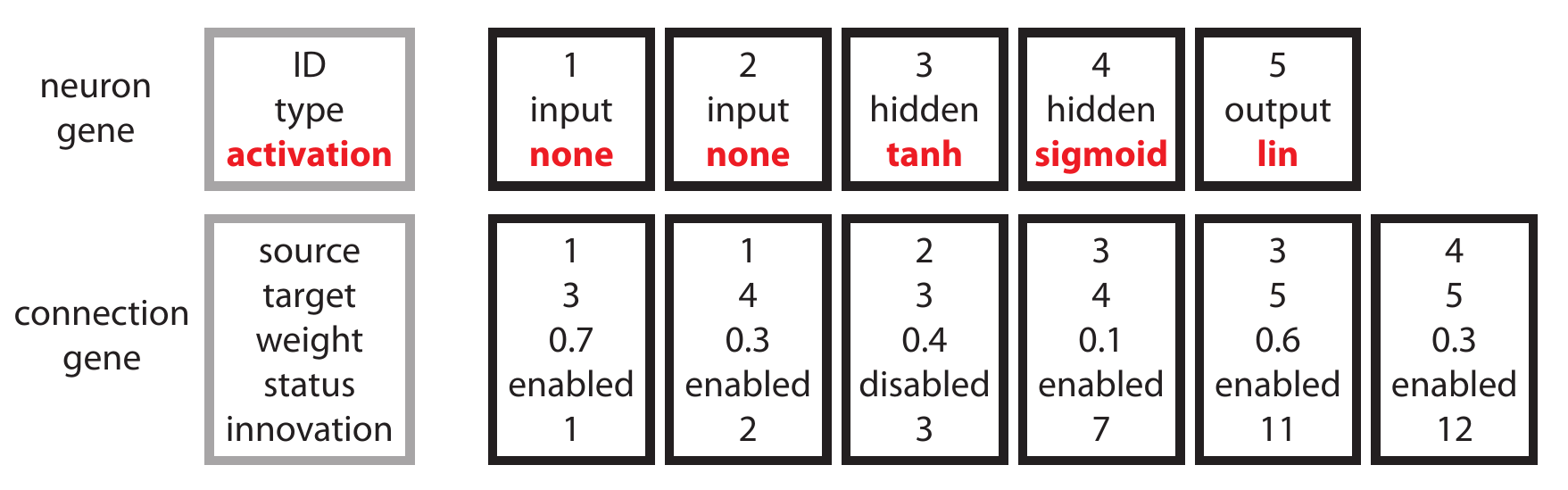}
\caption{Directly encoded HA-NEAT genome, sample input and output values are presented to the network directly.}
\label{fig:MAneatStructure} 
\end{figure}

\begin{figure}[ht!]
\centering
\includegraphics[width=\linewidth]{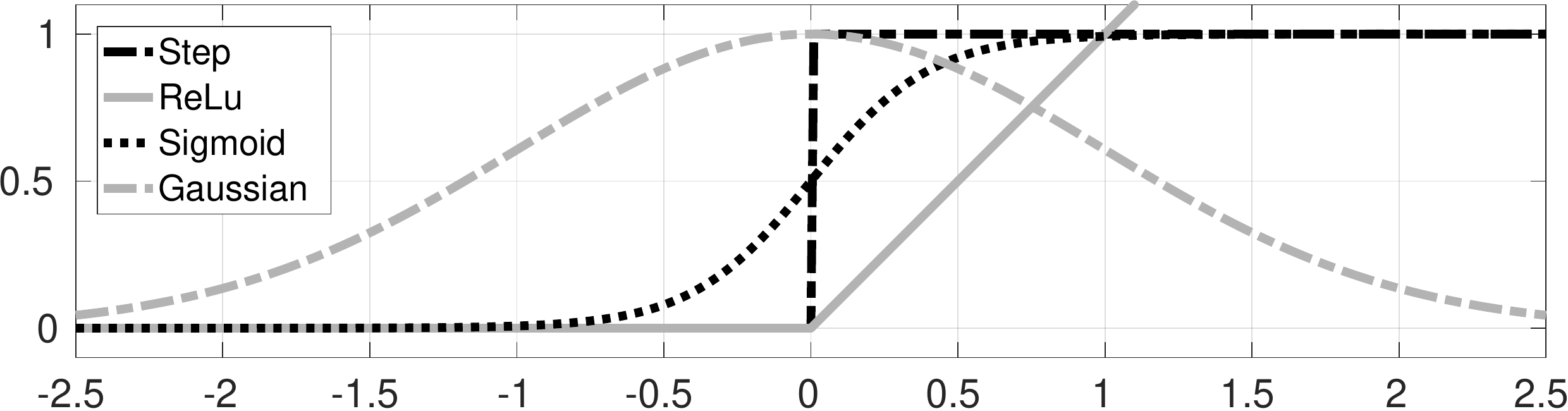}
\caption{Activation functions used in HA-NEAT}
\label{fig:activationfcns} 
\end{figure}

In order to evolve the activation function, the genome is extended by adding genes that represent the activation function of the neurons. This modified genome is illustrated in Figure \ref{fig:MAneatStructure}. It is similar to the CPPNs used in HyperNEAT~\cite{Stanley2007}, but instead of taking the coordinates of weights as input, we use a direct encoding. The data is presented to inputs and outputs in a direct manner.
The activation function gene is an integer number, representing the index in a list of activation functions. The list contains a number of activation functions: the discontinuous step function, the non-differentiable Rectifier Linear Unit~\cite{Nair}, the smooth sigmoid and locally active Gaussian kernel, which are shown in Figure \ref{fig:activationfcns}. Instead of evolving a parameterized activation function as was done in~\cite{Mayer1993}, which would add continuous parameters to the search space, we manually determine a small number of qualitatively different activation functions to reduce the search problem for NEAT.

Activation functions like the hyperbolic tangent are not used, because these functions need a different range in their output normalization. Mixed, heterogeneous networks, would then need to scale all neuron outputs differently, depending on their output range.

During the initialization phase, we assign a standard linear activation function to the input and output nodes. 
A modified version of the \textit{add node} mutation operator is used. When constructing a new neuron, a random activation function is selected.

A new mutation operation that changes the activation function of a node is added as well, which was not used in the HyperNEAT algorithm. Depending on the probability of this particular mutation, a node is randomly selected and an activation function is uniformly randomly selected from the list of possible functions. Since the connection weights to this neuron were optimized for the old activation function, the network might perform worse than its parents.
Therefore, highly non-linear changes in the networks are expected during evolution, when the activation function is mutated. This often leads to a drop in fitness first and decreases the chance of an individual to survive. 

Speciation is used to protect innovation. The \textit{mutate activation} operator is adjusted by changing the identifier of the node and the innovation numbers of \textit{all} incoming and outgoing connections if the activation function is changed. This increases the individual's speciation distance profoundly, as a single mutation can affect many connections' innovation numbers. This is a desirable effect, because changing the activation function in many neural pathways causes large qualitative changes in the neural network's output function. 

Large changes to individuals are now protected by speciation. In order to prevent too many changes to an individual, we only allow changing one node per genome per generation. The effect of mutating the activation function of existing nodes, as opposed to fixing the activation when the node is created, as in HyperNEAT, is examined in more detail in Section \ref{sec:evaluation}, where we show that the mutation operator is beneficial to evolution.

The expected impact of the proposed extension can be summarized as follows. We expect smaller networks to be more expressive, as was shown in Section \ref{sec:intro}. We therefore expect an approximation accuracy that rivals homogeneous networks created with NEAT, especially for problems with multiple target functions. The decrease in the number of necessary nodes to approximate a certain target will lead to faster convergence, although this will be partially counteracted by the increased search space, which depends on the number of possible activation functions.

\section{Evaluation}
\label{sec:evaluation}
\subsection{Data Sets}
\begin{figure}[htbp]
\centering
\includegraphics[width=1\linewidth]{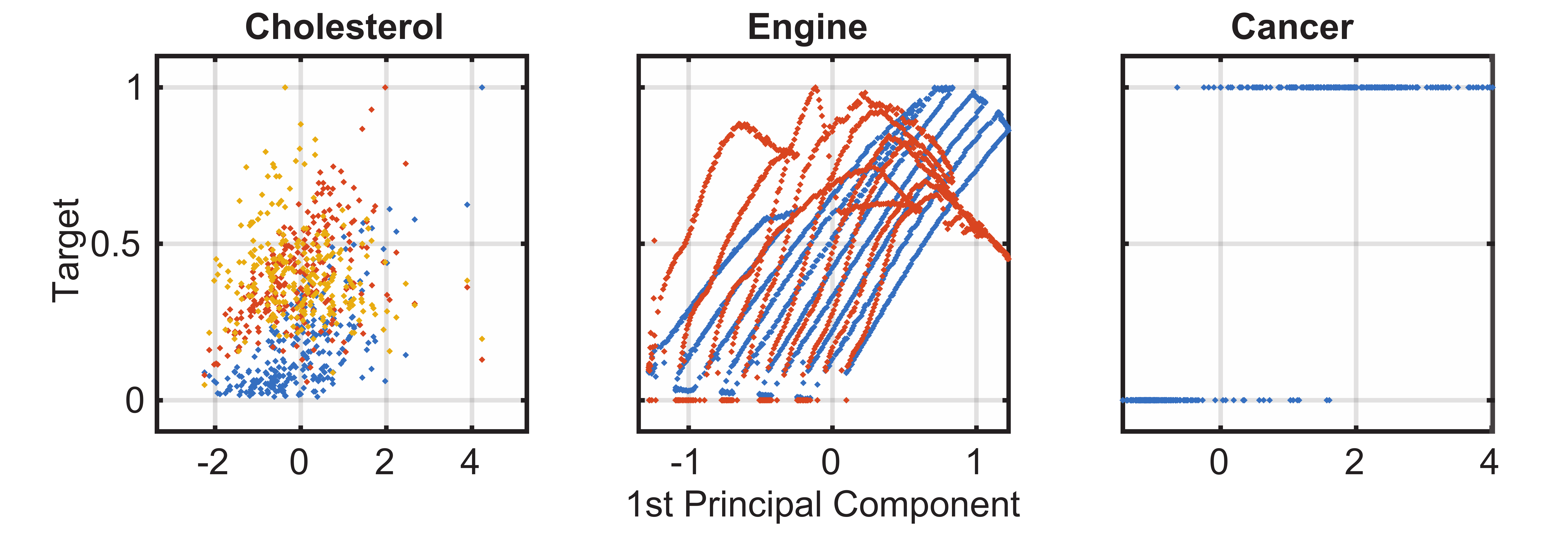}
\caption{Datasets, viewed on the first principal component, left: cholesterol levels prediction, middle: prediction of engine torque and CO2 emissions, right: breast cancer classification. Colors are used to indicate multiple target values.}
\label{fig:data} 
\end{figure}

A number of regression and classification datasets is used to evaluate HA-NEAT. All inputs are normalized to the interval $[-1,1]$. The datasets are separated into training and test sets using 5-fold cross validation, which are replicated 10 times, leading to a total of 50 replicates.
We compare the median mean square error (MSE), and 25\% and 75\% quantiles of the regression predictions with the target values, which are normalized between $[0,1]$.\\\\
\noindent\textbf{Cholesterol Level Indicators}\\
The first task is a regression problem in which three cholesterol level indicators (LDL, VLDL and HDL) from 21 spectral measurements of 264 blood samples are estimated. The data originates from work done by Purdie~\cite{Purdie1992} and is shown in Figure \ref{fig:data}. \\\\
\noindent\textbf{Engine Torque and Emissions}\\
The second task, also a regression problem, is to estimate torque and nitrous oxide emissions for 1199 samples containing fuel rate and speed. The data is provided by Martin Hagan\cite{Hagan2017}. \\\\
\noindent\textbf{Wisconsin Breast Cancer Diagnosis Problem}\\
The final task is a classification problem from the UCI~\cite{manga1990} machine learning database. The data set provides 699 samples, 16 of which are incomplete and therefore omitted for these experiments. Samples contain 9 dimensions, describing several known indicators for breast cancer and 2 classification labels: benign and malignant. The performance on this binary classification problem is measured by rounding the neural network's output to the closest label and comparing the MSE on the labels, so the classification threshold is set to 0.5 for this binary problem.

\subsection{Algorithmic Setup}

\begin{table}[h!tbp]
\centering                                                                      
\begin{tabular}{lcl}
\multicolumn{3}{c}{\textbf{General}}\\
\hline
population size & 100&\\
max. generation & 3000 & \\
\multicolumn{3}{c}{\textbf{Speciation}}\\
\hline
target species  & 10 &\\
compatibility threshold & 20 & start value\\
excess gene weight & 1 & \\
disjoint gene weight & 1 & \\
weight distance weight& 0.2 & \\
drop off age & 15 & \\
\multicolumn{3}{c}{\textbf{Genome operations}}\\
\hline
crossover& 90&\% of population\\
add node(*)& 1&\% chance/genome\\
add connection& 30&\% chance/genome\\
mutate activation(*)& 20 &\% chance/genome\\
\multicolumn{3}{c}{\textbf{Gene operations}}\\
\hline
mutate weight& 20&\% chance/gene\\
$\delta$ weight& 2&\\
enable connection& 0.02&\% chance/gene\\
disable connection& 0.2&\% chance/gene\\
\end{tabular}                                                                   
\caption{Hyperparameters of heterogeneous(*) NEAT}
\label{table:hyper}                                                      
\end{table} 

HA-NEAT's hyperparameters are described in Table \ref{table:hyper} and were selected based on experiments done with a simple regression problem which was not part of the data sets used in these experiments. A custom Matlab implementation of NEAT was used.
The data is normalized and used to train HA-NEAT and standard NEAT with a number of fixed activation functions.


\subsection{Homogeneous Networks}
As mentioned above, Kamruzzaman and Aziz~\cite{Kamruzzaman2002} and Laudani et al.~\cite{Laudani2015} show that there is no single best activation function for different regression problems. To confirm that the data sets indeed require different activation functions for optimal training, a performance comparison of homogeneous networks, networks that use a single activation function and evolved with NEAT, between the cholesterol and engine regression, as well as the cancer classification problems is made. 

\begin{figure}[h!tbp]
\centering
\includegraphics[width=1\linewidth]{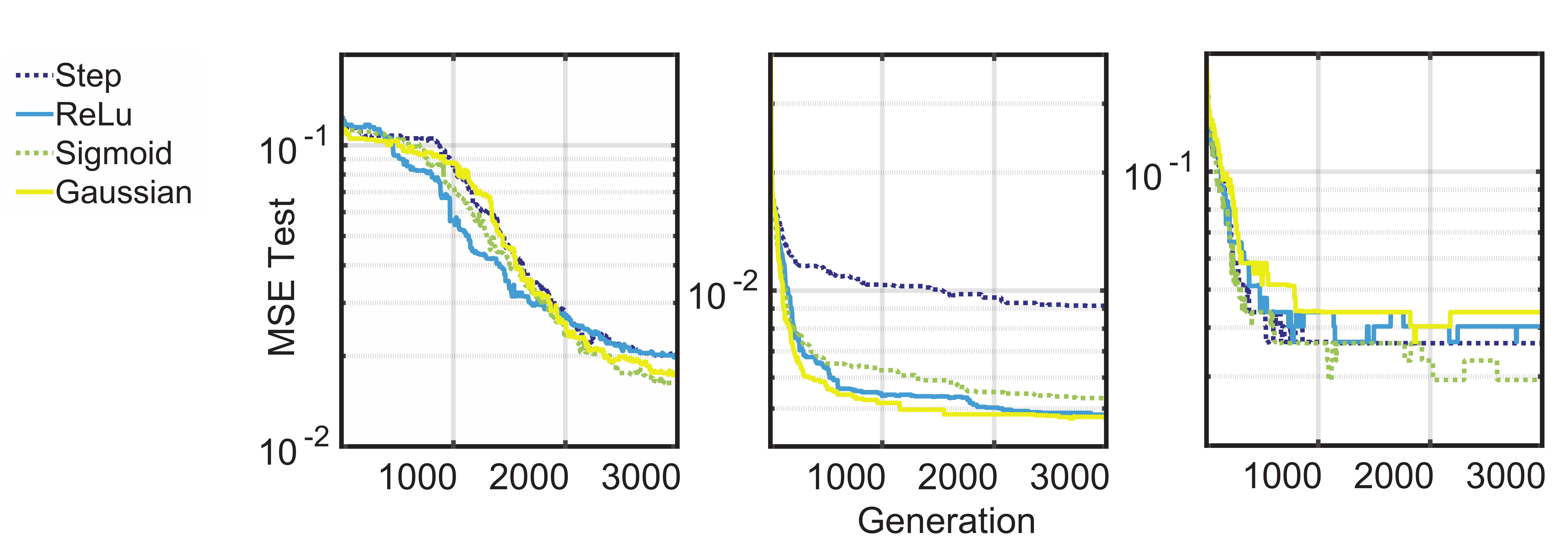}
\caption{Median test error over 50 replicates of homogeneous networks on the cholesterol, engine and cancer problem. Best homogeneous networks for the first set use a sigmoid activation, for the second set, Gaussian and ReLu have the highest accuracy, and the third set requires a sigmoid activation}
\label{fig:results:homogeneous:error} 
\end{figure}

\begin{table}[h!tbp]
\centering                                                                      
\begin{tabular}{c>{\columncolor[gray]{0.9}}ccccc}                                              
& HA-NEAT & Step & ReLu & Sigmoid & Gaussian\\
\hline
\multicolumn{6}{c}{\textbf{Cholesterol}}\\
MSE 		  &  0.0197    &0.0199    &0.0198   & 0.0163  &  0.0170\\ 
25\% quantile &    0.0173   & 0.0181   & 0.0167   & 0.0138  &  0.0149\\ 
75\% quantile &    0.0248    &0.0237   & 0.0251   & 0.0197 & 0.0214\\ 

\multicolumn{6}{c}{\textbf{Engine}}\\
MSE 		&  0.0048    &0.0091    &0.0047   & 0.0053   & 0.0048\\ 
25\% quantile &  0.0041   & 0.0083   & 0.0043   & 0.0047   & 0.0042\\ 
75\% quantile &  0.0055    &0.0103    &0.0054   & 0.0061   & 0.0058\\ 

\multicolumn{6}{c}{\textbf{Cancer}}\\
MSE 		&  0.0366   & 0.0365  &  0.0403  &  0.0294   & 0.0438\\ 
25\% quantile &  0.0292    &0.0292  &  0.0292   & 0.0219   & 0.0292\\ 
75\% quantile &  0.0438    &0.0441  &  0.0511  &  0.0511   & 0.0511\\ 

\end{tabular}                                                                   
\caption{Median MSE, 25\% and 75\% quantiles, comparing heterogeneous networks and homogeneous networks with different activation functions on cholesterol, engine and cancer data sets, 50 experiment replicates. HA-NEAT shows similar results on the cholesterol data set, although sigmoid creates the best performing networks. On the engine dataset, HA-NEAT performs as good as the best performing homogeneous networks, ReLu and Gaussian. The best performing homogeneous network for the cancer data set use sigmoid, heterogeneous networks' performance is not far off.}
\label{table:results:homogeneous:error}                                                      
\end{table} 

The results in Figure \ref{fig:results:homogeneous:error} show that indeed, not a single activation function leads to the best performing networks. Mean square errors are shown in Table \ref{table:results:homogeneous:error}. We conclude that no single activation function works best for our data sets, which confirms the results in~\cite{Turner2014}.

\subsection{Heterogeneous Networks}

\subsubsection{Mutation Operator}
\begin{figure}[h!tbp]
\centering
\includegraphics[width=1\linewidth]{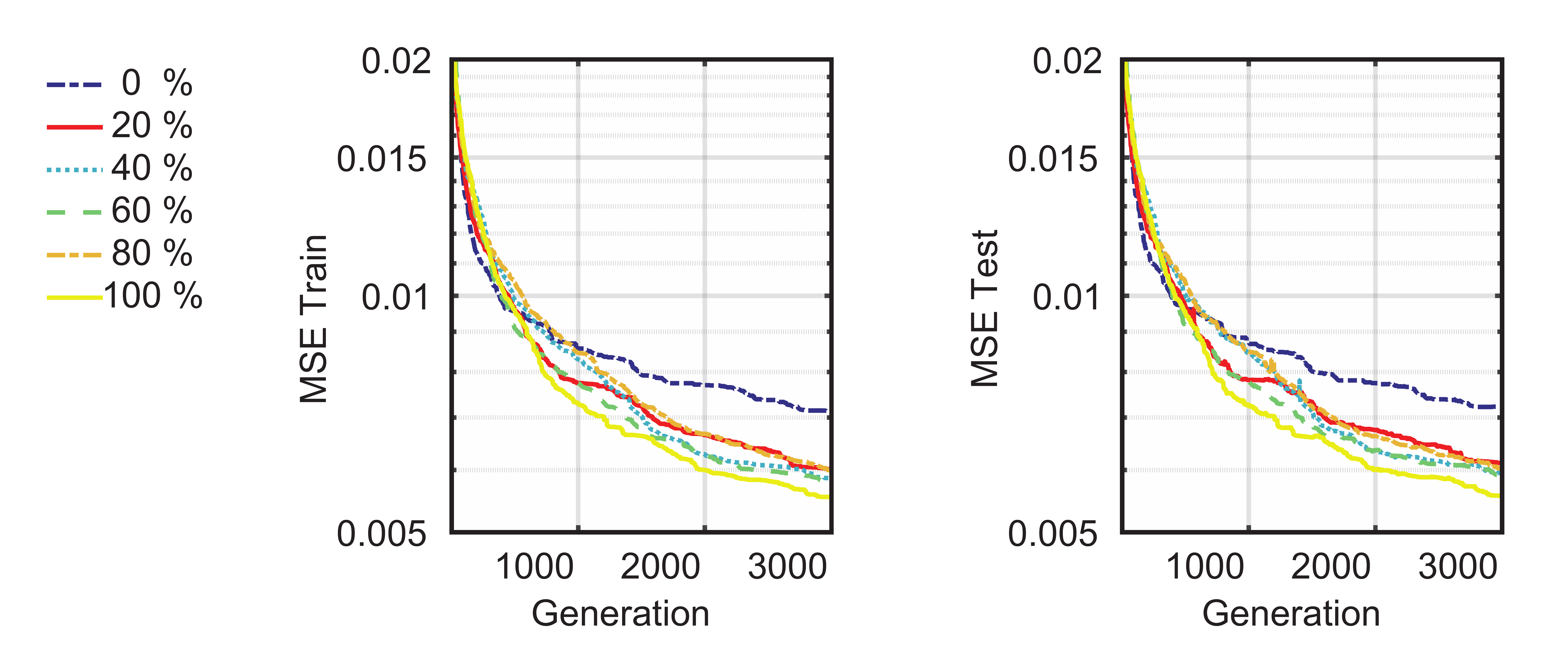}
\caption{Mean training and test error for various rates of mutating the activation function once per individual genome on the engine data set.}
\label{fig:results:mutact} 
\end{figure}

The mutation function that changes the activation function of a node is an addition that was not used in the HyperNEAT algorithm. Figure \ref{fig:results:mutact} shows the difference between not using the mutation (0\% mutation rate) and various probabilities of mutating once per genome  on the engine data set. A rate of 100\% means that \textit{one of the nodes} in an individual's network is changed in every generation. Experiments were repeated 50 times but only for a small population of 50 individuals, explaining the discrepancy with respect to results of other experiments.

Although the algorithm converges quicker in the beginning when the mutation operator is disabled (similar to what is done in HyperNEAT), after 500-750 generations, depending on the mutation rate, all instances of HA-NEAT surpass the accuracy of the mutation-free instance. Although the operator can be quite destructive, it still increases the evolvability of our representation. The difference between various mutation rates seems to not be very significant, but this needs to be analyzed more in depth.

\subsubsection{Accuracy and Size}

\begin{figure}[h!tbp]
\centering
\includegraphics[width=1\linewidth]{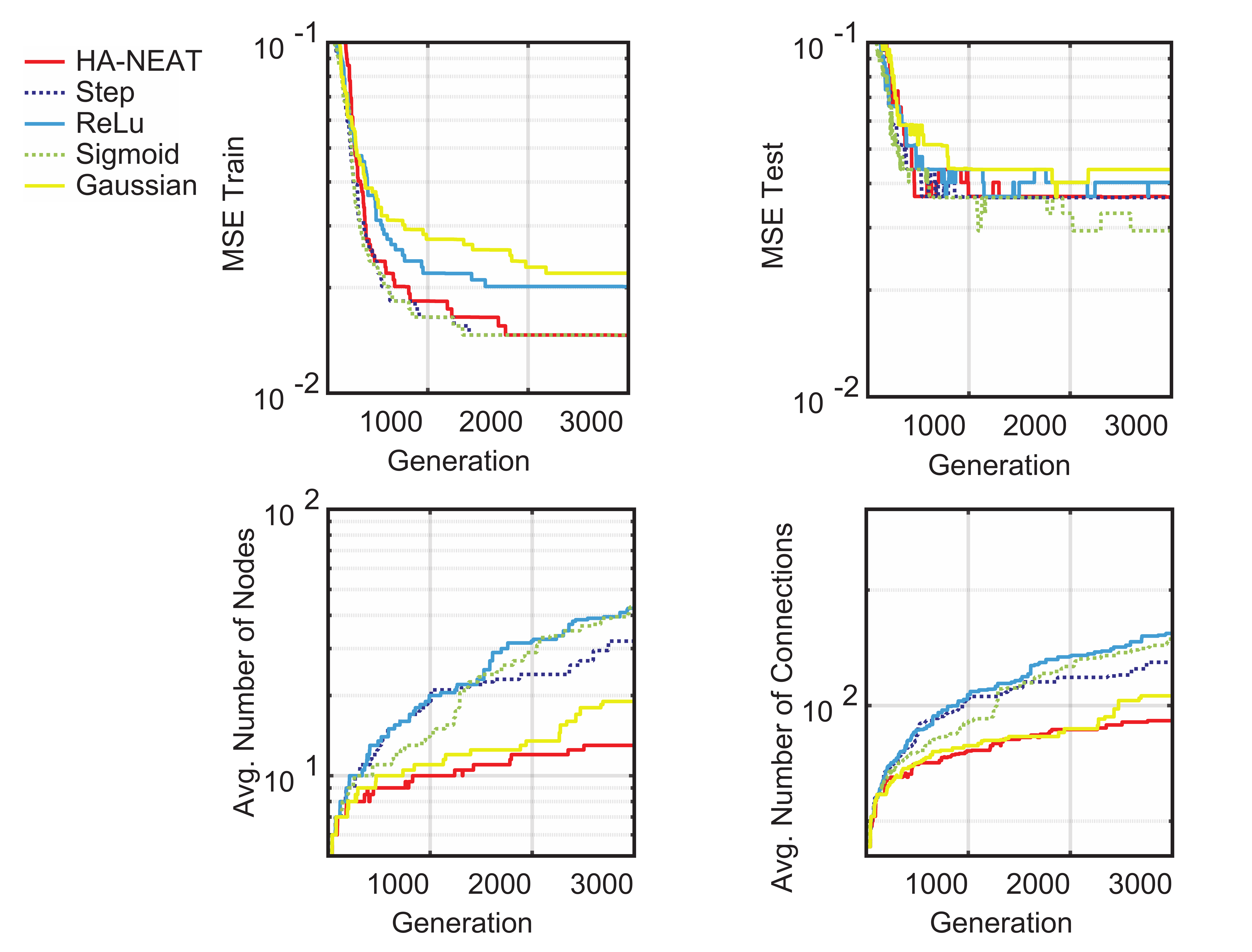}
\caption{Development of median training and test errors, and network size (nodes and connections) compared for engine dataset. 
HA-NEAT's training and test error are as good as any homogeneous network. The number of nodes and connections is smaller than the best performing homogeneous network.}
\label{fig:results:heho:devel} 
\end{figure}

Figure \ref{fig:results:heho:devel} shows the development of median training error, test error and number of nodes and connections on the cancer dataset over 50 replicates. The heterogeneous networks perform as well as the best homogeneous networks, but with lower complexity.

\subsubsection{Activation Functions}
\begin{figure}[h!tbp]
\centering
\includegraphics[width=1\linewidth]{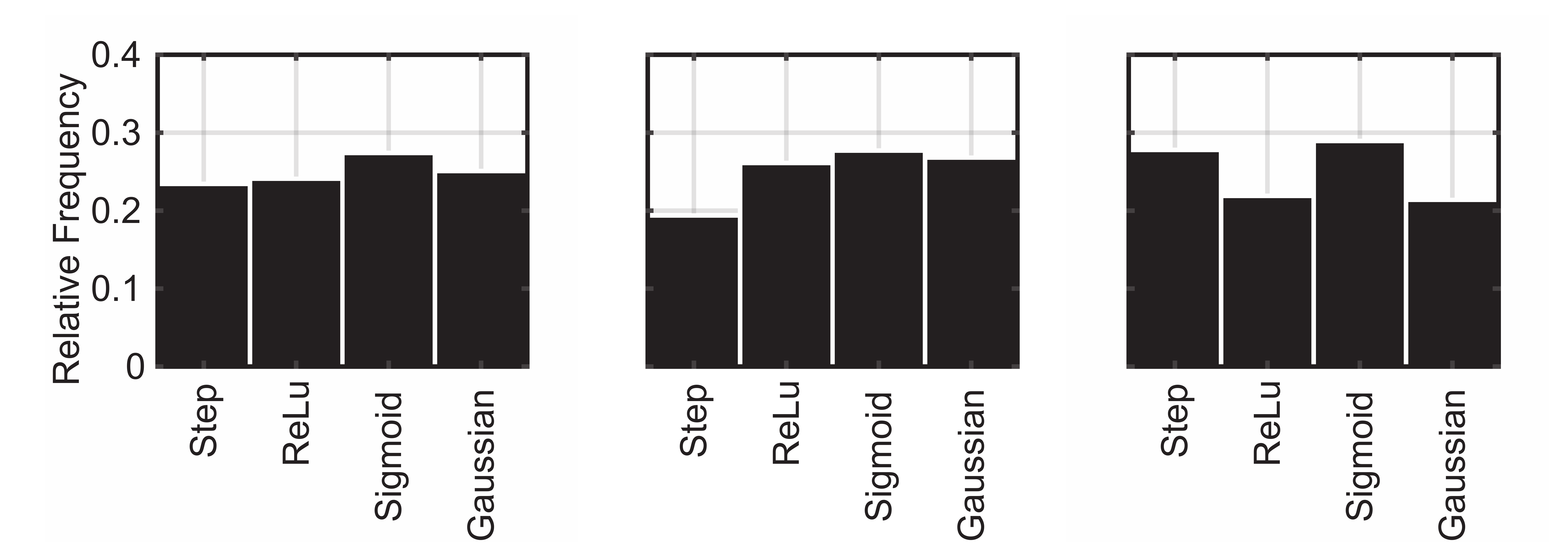}
\caption{Relative frequency of activation functions in resulting heterogeneous networks for cholesterol, engine and cancer datasets.}
\label{fig:results:heho:actfcns} 
\end{figure}

The median amount of nodes used in the heterogeneous networks for the three data sets, 32/10/20, is much lower than the 51/16/29 nodes required in the homogeneous networks. Similarly, the heterogeneous networks use a median value of 124/35/116 connections, whereas the homogeneous networks use 225/170/197 connections. This shows that HA-NEAT creates more parsimonious networks than the original implementation of NEAT. Although the topological search space, in terms of different neurons, is increased, because of the added dimension of 4 instead of 1 activation function, the algorithm has a convergence rate similar to that of the homogeneous versions.

The relative frequency of different activation functions in the heterogeneous solutions is shown in Figure \ref{fig:results:heho:actfcns}. For the cholesterol HA-NEAT does not seem to prefer a specific activation function. In case of the engine data, the step function is underrepresented, whereas in case of cancer classification, step and sigmoid activations are preferred. The histograms show that HA-NEAT does not simply prefer just finding the best performing homogeneous network, but finds qualitatively different solutions.

\subsubsection{Parsimony}

\begin{figure*}
\centering
\includegraphics[width=1\linewidth]{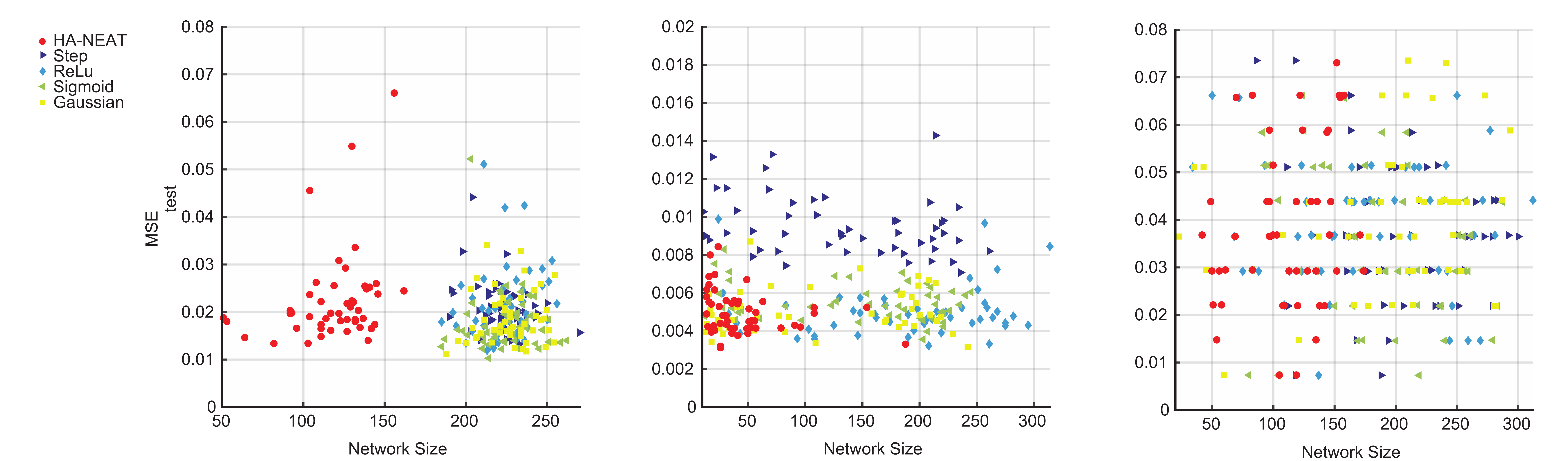}
\caption{Network size and test errors compared between heterogeneous and homogeneous networks for the cholesterol (left), engine (middle) and cancer (right) data sets.}
\label{fig:results:networksize} 
\end{figure*}

\begin{figure*}[b]
\centering
\includegraphics[width=1\linewidth]{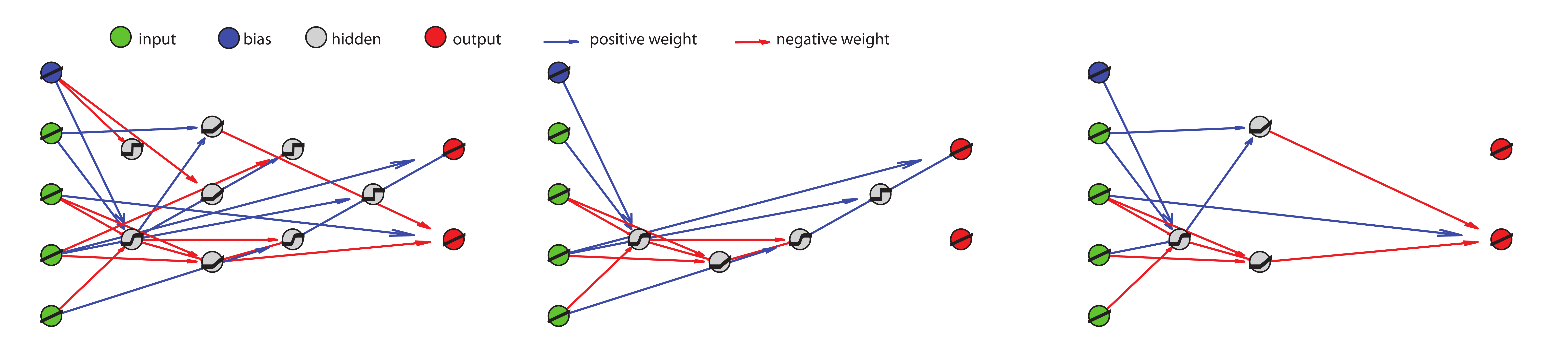}
\caption{One of the best networks found for engine data set. Left: full network, middle: subnetwork serving the first output node, right: subnetwork serving the second output node.}
\label{fig:results:modularity} 
\end{figure*}

Figure \ref{fig:results:networksize} shows the relationship between test errors in a comparison between heterogeneous and homogeneous networks. When trained on the cholesterol data set NEAT is clearly more parsimonous when creating heterogeneous instead of homogeneous networks, showing that it is easier to be more parsimonious when we allow the evolution of activation functions. As was already discussed in Figure \ref{fig:results:heho:devel}, although the search space is increased by the number of activation functions, the increased expressivity of the networks decreases it with respect to a certain accuracy. Similar results are noticeable for the engine data set. Here, the resulting variance in terms of accuracy and size is much smaller using heterogeneous networks, as they have a much smaller footprint in the plot. In the classification experiment the heterogeneous networks are again much smaller than most homogeneous networks.

Although we did not focus on the topic of modularity in this work, Figure \ref{fig:results:modularity} shows one of the top networks produced by HA-NEAT on the engine data set. It shows that for this multi-target problem, we do see the kind of modularity we describe in \ref{sec:intro:heterogeneous}. The two hidden neurons in the lower left of the network are used for both outputs, whereas the rest is split into two modules that serve only one of the outputs. This is the same behaviour as was observed for HyperNEAT in~\cite{Huizinga2014}.

In Section \ref{sec:related:activation} it is mentioned that Turner and Miller~\cite{Turner2014} found that heterogeneous networks perform better than the average homogeneous networks. We show that for regression and classification problems heterogeneous networks perform not only as well as the average homogeneous model, but mostly as well as the \textit{best} homogeneous models. Most of all though, the heterogeneous networks are more parsimonious and do indeed contain a non-trivial mix of selected activation functions. The networks reach a higher accuracy per connection, resulting in a smaller search space. 

\section{Conclusion}
\label{sec:conclusion}
An extension to the NEAT algorithm is introduced, allowing for it to produce heterogeneous networks, consisting of neurons with different activation functions, a directly encoded version of the CPPNs used in HyperNEAT. HA-NEAT is able to find solutions that perform as good as the best homogeneous networks, which use only one kind of activation function, but using less nodes and connections. Although the search space is increased by parameterizing the activation function of every node, the reduction in topological complexity is shown to lead to convergence speeds that are similar to the best-performing homogeneous networks. HA-NEAT automatically picks solutions that rivals the original NEAT in regression problems, posing the benefit of not having to pick the right activation function and at the same time being more parsimonious than homogeneous networks.

A mutation operator that allows changing the activation function of nodes during evolution has been evaluated. This operator was not used in HyperNEAT but shows that it improves convergence. The amount of activation mutations did not have a significant impact but the results seem to suggest, that we may use more than one mutation per genome per generation. This will have to be further analyzed.

Although heterogeneous networks perform as well as the best homogeneous networks, more attention has to be given to the training of weights. Using backpropagation on the evolved network topologies in certain intervals might allow us to make fairer comparisons between individuals with different topologies sooner, which would decrease some of the necessity for innovation protection. 

When training data is scarce and overfitting is to be expected, networks should be as parsimonious as possible, while still retaining a high accuracy. The use of heterogeneous networks has shown to be a promising approach to do just that in topology-constructing neuroevolutionary training methods. 

\bibliographystyle{ACM-Reference-Format}
\bibliography{HeteroNEAT} 

\end{document}